\def\year{2022}\relax
\documentclass[letterpaper]{article} 
\usepackage{aaai22}  
\usepackage{times}  
\usepackage{helvet}  
\usepackage{courier}  
\usepackage[hyphens]{url}  
\usepackage{graphicx} 
\usepackage{natbib}  
\usepackage{caption} 
\DeclareCaptionStyle{ruled}{labelfont=normalfont,labelsep=colon,strut=off} 
\usepackage{algorithm}
\usepackage{algorithmic}

\usepackage{newfloat}
\usepackage{listings}
\floatstyle{ruled}
\newfloat{listing}{tb}{lst}{}
\floatname{listing}{Listing}

\usepackage{latexsym}
\usepackage{graphicx}
\usepackage{subcaption}
\usepackage{xcolor}
\usepackage{bm}
\usepackage{amsfonts}
\usepackage{amsmath}
\usepackage{booktabs}
\usepackage{multirow}
\usepackage{enumitem}

\title{Cross-Lingual Text Classification with Multilingual Distillation and Zero-Shot-Aware Training}
\author{
Ziqing Yang$^\dag$,
Yiming Cui$^\ddag$$^\dag$,
Zhigang Chen$^\dag$,
Shijin Wang$^\dag$
}
\affiliations{
$^\dag$State Key Laboratory of Cognitive Intelligence, iFLYTEK Research, China \\
$^\ddag$Research Center for SCIR, Harbin Institute of Technology, Harbin, China \\

$^\dag$\tt\{zqyang5,ymcui,zgchen,sjwang3\}@iflytek.com \\
$^\ddag$\tt ymcui@ir.hit.edu.cn
}

\begin{document}
\maketitle
\begin{abstract}

Multilingual pre-trained language models (MPLMs) not only can handle tasks in different languages but also exhibit surprising zero-shot cross-lingual transferability. However, MPLMs usually are not able to achieve comparable supervised performance on rich-resource languages compared to the state-of-the-art monolingual pre-trained models. In this paper, we aim to improve the multilingual model's supervised and zero-shot performance simultaneously only with the resources from supervised languages. Our approach is based on transferring knowledge from high-performance monolingual models with a teacher-student framework. We let the multilingual model learn from multiple monolingual models simultaneously. To exploit the model's cross-lingual transferability, we propose MBLM (multi-branch multilingual language model), a model built on the MPLMs with multiple language branches. Each branch is a stack of transformers. MBLM is trained with the zero-shot-aware training strategy that encourages the model to learn from the mixture of zero-shot representations from all the branches. The results on two cross-lingual classification tasks show that, with only the task's supervised data used, our method improves both the supervised and zero-shot performance of MPLMs.

\end{abstract}

\section{Introduction}

With the help of self-supervised objectives and large amounts of corpora, large pre-trained language models \cite{devlin-etal-2019-bert, liu2019roberta} have achieved state-of-the-art results and prevailed in NLP tasks. But considering that there are thousands of languages in the world, the monolingual pre-trained language models are only available for a small number of languages that have rich resources.
Multilingual pre-trained language models ({MPLM}s), such as multilingual BERT (mBERT), XLM \cite{DBLP:conf/nips/ConneauL19} and XLM-R \cite{conneau-etal-2020-unsupervised}, extend the monolingual pre-trained models to a hundred languages by training on multilingual corpora. MPLMs have learned cross-lingual language representations and exhibited surprising zero-shot cross-lingual effectiveness \cite{wu-dredze-2019-beto, pires-etal-2019-multilingual}. When fine-tuned on a downstream task in a source language, MPLMs can generalize to different target languages without seeing parallel task data (aligned pairs of sentences in a pair of languages). 
MPLMs have greatly facilitated building multilingual NLP pipelines by reducing the effort of pre-training different models for different languages.  

However, cross-lingual transferability comes at a price. 
Even though it has been trained on an enormous amount of corpora, a model like XLM-R still performs worse on rich-resource languages than the monolingual models that have been trained on a comparable amount of data with a comparable model size \cite{conneau-etal-2020-unsupervised}. 
One possible explanation is the limit of model capacity \cite{conneau-etal-2020-unsupervised, wang-etal-2020-negative}.

Is it possible to make the multilingual model perform well on supervised languages and meanwhile retain a good cross-lingual transferability after fine-tuning? 
In this work, we try to address this issue from two perspectives: 
(1) leveraging monolingual models by knowledge transfer; (2) exploiting the internal cross-lingual transferability of MPLMs with a new model structure and new training strategy.

\begin{figure}[tbp]
    \centering
    \includegraphics[width=.75\columnwidth]{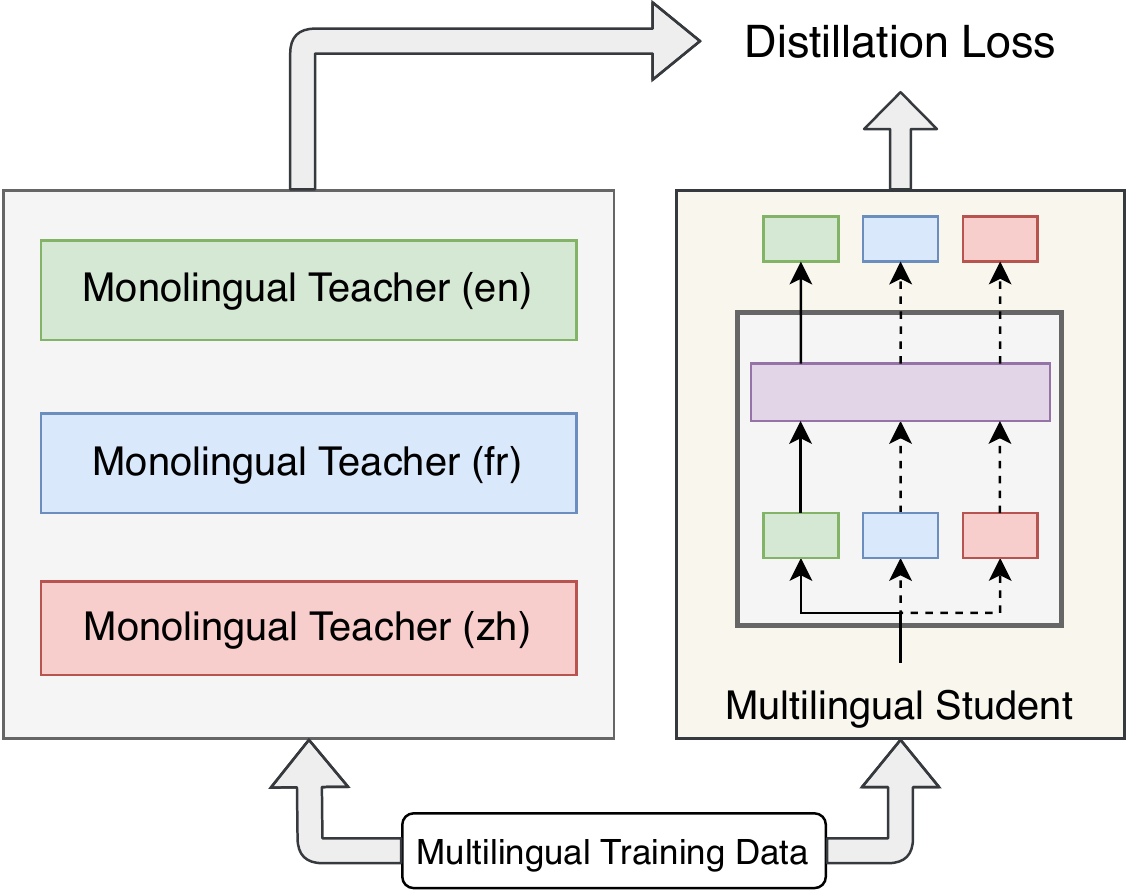}
    \caption{The overview of multilingual knowledge distillation with MBLM as the student model.}
    \label{fig:overall}
\end{figure}

To utilize the knowledge from high-performance monolingual pre-trained language models, we propose a distillation-based framework.
To note that multilingual fine-tuning, i.e., training the multilingual model on multiple language datasets, brings more improvements than training on a monolingual dataset \cite{conneau-etal-2020-unsupervised, huang-etal-2019-unicoder}, we bring this idea into the distillation. We find that by letting the multilingual student learn from multiple monolingual teachers simultaneously, the overall performance can be improved, especially on supervised languages.

Another important issue of MPLMs is the existence of gradient conflicts and language-specific parameters \cite{wang-etal-2020-negative}, which means that different languages are fighting for model capacity and optimizing the model towards different directions. 
To take the issue into consideration and exploit the internal cross-lingual transferability of MPLMs, we propose a \textbf{M}ulti-\textbf{B}ranch multilingual \textbf{L}anguage \textbf{M}odel structure, namely MBLM. It is built on the MPLM such as XLM-R and mBERT.
Compared to XLM-R and mBERT, MBLM replaces the transformer blocks at the bottom of the encoder with language-specific branches for each supervised language to allocate more capacity.
During training, we introduce an interaction mechanism between language branches. Intuitively, we let each branch be aware of the zero-shot representation from other branches. We call this strategy \textit{zero-shot-aware training}. It exploits and further boosts the cross-lingual transferability of  MPLMs.

Our method is computationally inexpensive.  It only uses task data and needs no additional corpora or pre-training procedures. Besides, it does not require parallel training data between different languages, so it is easy to be applied for different languages and different tasks.  We test our method on two cross-lingual classification tasks, the model performance is consistently improved. To summarize, our contributions are:
\begin{itemize}[leftmargin=*,noitemsep]
\item We propose to use the multilingual knowledge distillation with multiple monolingual teachers to improve the model's downstream task performance. 
\item We propose an MBLM structure and its zero-shot-aware training strategy which exploits the cross-lingual transferability more effectively.
\item We conduct experiments on two cross-lingual classification tasks. The results outperform baselines in both supervised and zero-shot performance.
\end{itemize}

\section{Related Work}
\noindent \textbf{Multilingual Pre-trained Language Models}. Besides Masked Language Modeling (MLM) \cite{devlin-etal-2019-bert} and Translation Language Modeling (TLM) \cite{DBLP:conf/nips/ConneauL19}, various kinds of multilingual pre-training objectives have been proposed.
Unicoder \cite{huang-etal-2019-unicoder} trains the model with new cross-lingual pre-training tasks, such as cross-lingual word recovery, cross-lingual paraphrase classification, and cross-lingual MLM. 
InfoXLM \cite{chi2020infoxlm} proposed a pre-training task based on contrastive learning, from an information-theoretic perspective.
\citet{DBLP:journals/corr/abs-2010-12547} also introduced an alignment method based on contrastive learning.
\citet{DBLP:conf/iclr/CaoKK20} proposed an explicit word-level alignment procedure. 
All the mentioned multilingual pre-training objectives have leveraged parallel data that were not used int training mBERT and XLM-R.

\noindent \textbf{Downstream Tasks Fine-tuning}. To increase the multilingual model performance on downstream tasks without introducing another costly pre-training stage that trains the whole MPLMs, special model architectures have been proposed. Filter model \cite{DBLP:journals/corr/abs-2009-05166} encodes text input and its translation simultaneously and then performs cross-language fusion to extract multilingual knowledge.  Another stream of works is adding \textit{adapters} \cite{DBLP:conf/icml/HoulsbyGJMLGAG19, pfeiffer-etal-2020-mad,DBLP:journals/corr/abs-2012-06460} to the MPLMs. The newly added adapters need to be trained with MLM then fine-tuned on the downstream tasks.
Compared with the existing pre-training or fine-tuning methods, which rely on extra corpora, in this work, we only use task training sets for training.

\noindent \textbf{Cross-lingual Knowledge Transfer}. 
Knowledge distillation \cite{DBLP:journals/corr/HintonVD15} is a classical and effective method to transfer knowledge from a teacher model to a student model. 
\citet {xu-yang-2017-cross} uses knowledge distillation to transfer between two text classification models in different languages with the help of parallel task data and unlabeled task data. 
\citet{chi-etal-2020-monolingual} investigates transferring knowledge from a monolingual model to a multilingual model. 
\citet{textbrewer-acl2020-demo} finds that distillation from multiple teachers outperforms simple ensemble models.
\citet{wu-etal-2020-single} distills from multiple multilingual models to a single multilingual model on NER tasks.
Knowledge can also be transferred without using distillation. \citet{artetxe-etal-2020-cross} transfers a monolingual language model to a new language by learning the new language embedding matrix through MLM.

\section{Methodology}

\subsection{Tasks and Settings}

We consider a multilingual task covering $N_{L}$ languages. The languages are divided into two sets: A \emph{supervised} set $L_{sp}$ contains $N_{sp}$ languages and a \emph{zero-shot} set $L_{zs}$ contains $N_{zs}$ languages.
Both the training and evaluation sets are available for the supervised languages. While for the zero-shot languages, there are no training datasets. We only perform the zero-shot evaluation on the zero-shot languages.
We do not assume the parallelism of the training sets from different supervised languages.

Our multilingual model is to be built on MPLM. Besides, we also have large monolingual pre-trained language models for each supervised language. The goal is to obtain a multilingual model that performs well on both supervised languages and zero-shot languages while only with resources from supervised languages.
In this work, we evaluate our method on text classification tasks. 

The overview of the framework is shown in Fig \ref{fig:overall}. The framework consists of multilingual knowledge distillation and a novel network structure for the student model, which will be explained below.

\subsection{Multilingual Knowledge Distillation}
Knowledge distillation is an effective method to improve the model supervised performance.
To transfer knowledge from multiple monolingual teachers in different languages to a multilingual student model, we perform knowledge distillation in a multi-task way. We call this process {\textbf{MultiKD}}.

For a multilingual task that contains training sets $\{\mathcal{D}_1, \mathcal{D}_2,\ldots, \mathcal{D}_{sp}\}$,
we first fine-tune the monolingual teacher models $\{\mathcal{T}_1,\mathcal{T}_2,\ldots\ \mathcal{T}_{sp}\}$ on the corresponding training sets.
Then we train the student with knowledge distillation. In each iteration, we randomly select a language $l$ from the supervised languages set, then pick the corresponding dataset $\mathcal{D}_l$ and sample a batch of training examples $\mathcal{B}=\{\bm{x}^{(l)}_{i},y^{(l)}_{i}\}$. We feed the batch $\mathcal{B}$ to both the monolingual teacher $\mathcal{T}_{l}$ and the student $\mathcal{S}$, which return  the teacher logits $\bm{z}^{(T_l)}_i$ and the student logits $\bm{z}^{(S)}_i$ ($i\in \{1,\ldots,N\}$ ) respectively. The  classification probabilities given by the teacher and the student are calculated by
\begin{align}
&\bm{p}^{(T_l)}_i = \text{softmax} (\bm{z}^{(T_l)} / \tau) \\
& \bm{p}^{(S)}_i = \text{softmax} (\bm{z}^{(S)} / \tau)
\end{align}
where $\tau$ is the distillation temperature. The distillation loss is the cross-entropy loss between the two probabilities
\begin{align}
\mathcal{L}^{(l)}_{\text{KD}}(\mathcal{B}, \theta_{S}) = -\frac{1}{N}\sum_{i}^{N}  \bm{p}^{(T_l)}_i \cdot \log \bm{p}^{(S)}_i
\end{align}
 where $N$ is the batch size, $\theta_S$ denotes the parameters of the student model.
We optimize the student model by performing one step of gradient descent and continue to the next iteration.

As we repeat the above process, the student model continuously learns from different monolingual teachers.  The process is effectively equivalent to minimizing the following multilingual loss 
\begin{align}
\mathcal{L}_{\text{KD}}(\mathcal{B}, \theta_{S}) =\sum_{l}^{N_{sp}} \mathcal{L}^{(l)}_{\text{KD}}(\mathcal{B}, \theta_{S}) 
\end{align}

\subsection{Student Model}
Fine-tuning a single model with multilingual examples is beneficial in most cases. However, some works have found that this approach may lead to performance degradation \cite{mueller-etal-2020-sources}.
Further studies \cite{wang-etal-2020-negative} show that there exist language-specific parameters which lead to the competition for capacity between different languages.
Inspired by the recent works, we propose an multi-branch model as our student model that allocates additional capacity for the supervised languages to relieve the competition between supervised and zero-shot languages.
During training, the model is optimized with a zero-shot-aware mechanism that takes advantage of multilinguality.
We describe our method in detail in the following.

\subsubsection{Pre-trained Language Model}
The MPLM such as mBERT and XLM-R consists of an embedding layer, followed by transformer blocks and a task-specific prediction head.  
Given an input sequence 
$\bm{x} = [w_1,w_2,\ldots,w_n]$  of length $n$, the embedding layer encodes $\bm{x}$ as 
\begin{equation}
H_0 = \textrm{Embedding}(\bm{x},\bm{f})
\end{equation}
where $H_0 \in \mathbb{R}^{n\times d}$ is the representation of the sequence. $\bm{f}$ refers to the optional features that may be fed to the embedding layer, such as segment ids or language ids.
 The transformer blocks further encode the representation successively as
\begin{equation}
H_i = \textrm{Transformer}_{i}(H_{i-1}),~~i\in\left[1,\ldots,L\right]
\end{equation}
For the classification task, we take the vector at the first position of the last hidden representation $H_L$ and feed it to the single-layer classification head. The outputs from the head are the logits, which give the probability for each class label by taking the softmax.

\subsubsection{Multi-Branch Multilingual Language Model (MBLM)}

In the text classification tasks, the predictions are class labels, which can stand for entailment relation, question-answer relation, sentiment polarity, etc.
The semantic meanings of class labels are independent of the languages, i.e., the model predictions are less language-specific.
This is opposite to the situations in structured prediction tasks, such as POS tagging and named entity recognition, where the predictions are highly dependent on the language of the input.

\begin{figure}[tbp]
\centering
\includegraphics[width=.7\columnwidth]{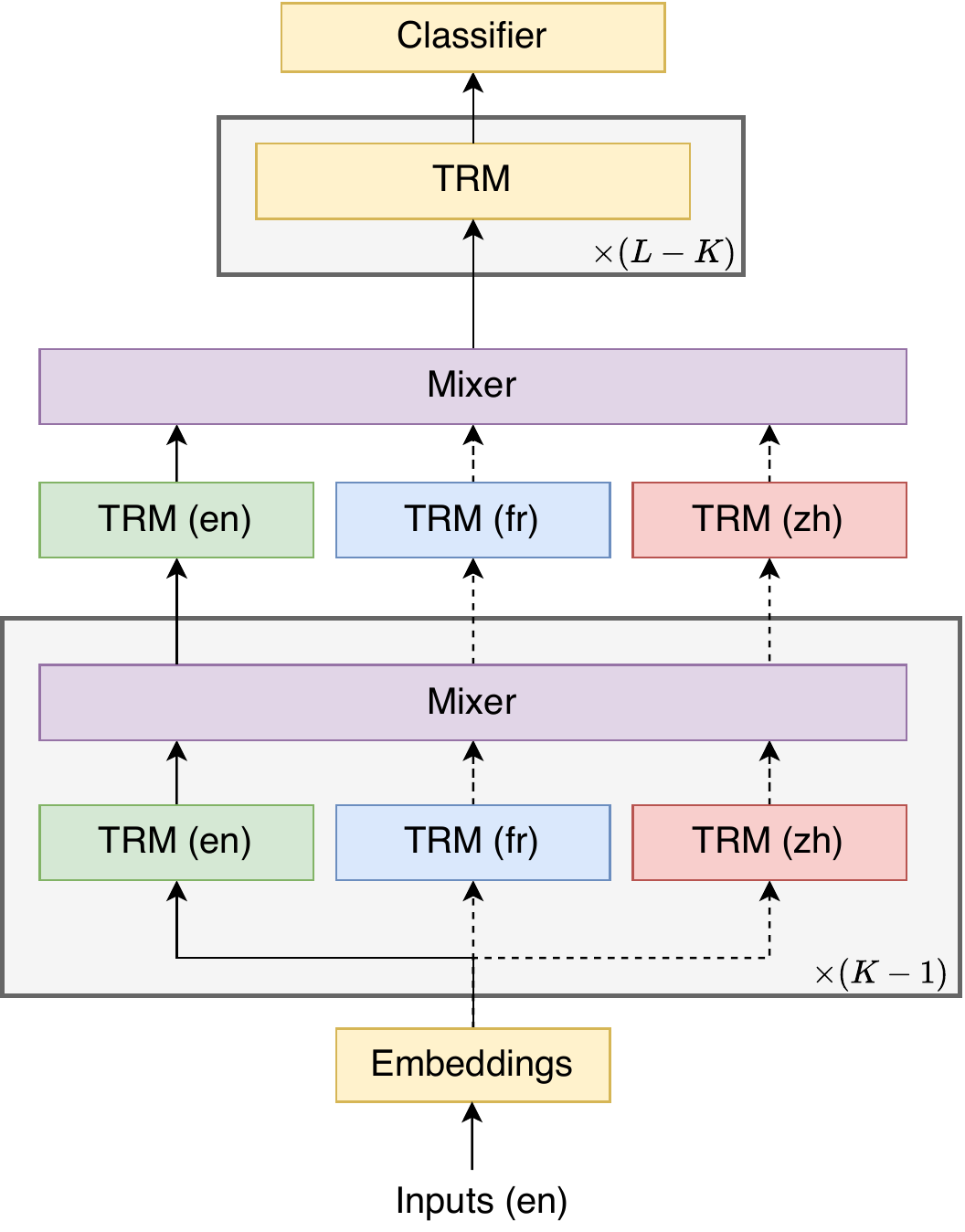}
\caption{The MBLM. We illustrate the structure with three language branches (en, fr, zh) and English inputs. The arrows denotes the forward pass directions. 
The backward pass can go through the solid lines, while it cannot go through dashed lines. {TRM} is short for Transformer.}
\label{fig:hydra}
\end{figure}

Thus we expect that when fine-tuning an MPLM on text classification tasks,  the representations in the upper layers of transformer blocks are encoded with more high-level semantic features and are more language-independent;
while as the bottom layers are closer to the embedding layer, they are processing low-level language-specific features.
Following this idea, we replicate the bottom $K$ layers of transformer blocks multiple times and construct $N_{sp}$ branches, where $N_{sp}$ is the number of supervised languages, as shown in Fig \ref{fig:hydra}. Each branch deals with its specific supervised language. The rest of  $L-K$ layers are shared between different languages and stay unchanged. 

\begin{figure}[tbp]
    \centering
    \includegraphics[width=0.7\columnwidth]{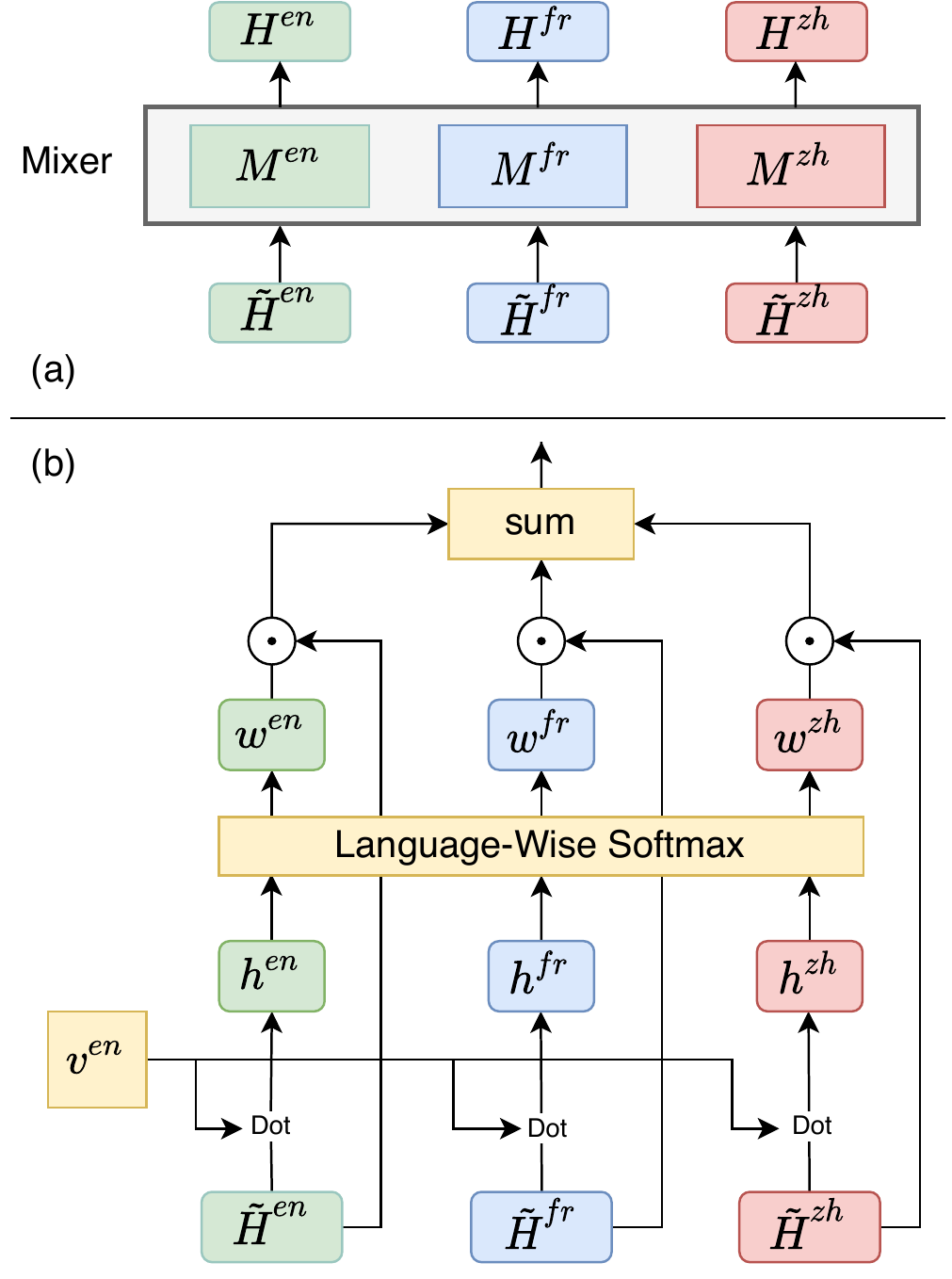}
    \caption{We illustrate the mixer module with three language branches. (a) A mixer includes several submodules $M^{en}$, $M^{fr}$ and $M^{zh}$. (b) The internal structure of the submodule $M^{en}$.}
    \label{fig:mixer}
    \end{figure}

\subsubsection*{$\bullet$~~Zero-Shot-Aware Training}

By adding branches for each supervised language, the model gets more capacity. However, the drawback is that we lose the multilinguality of the bottom encoders. 
Also, since each language-specific branch only sees the monolingual training data in its specific language, the multilinguality of full training data is not fully utilized.
To bring the advantage of multilingual training back and improve the zero-shot ability, we propose to mix the zero-shot representations from other branches during training and inference.
The model structure is shown in Fig \ref{fig:hydra}. In addition to adding language-specific branches, we mix their representations after each layer.

Suppose the current batch is in language $k$. During training and inference, we feed the batch to all the branches. At the $i$-th layer ($i\le K$), we have hidden representations from the transformers
\begin{equation}
\tilde H^{(l)}_{i} = \textrm{Transfomer}^{(l)}_{i}(H^{(l)}_{i-1})
\end{equation}
where $H^{(l)}_{0} = H_{0}$ for any supervised language $l$.

To make the each branch $l$ aware of the zero-shot representations from other branches $l'\neq l$, we mix the representations then feed it to the next layer. But before doing that, 
we detach  $\tilde H^{(l)}_{i}$ from the computation graph  if $l\neq k$:
\begin{equation}\label{eq:detach}
\tilde H^{(l)}_{i} \leftarrow \texttt{Detach} (\tilde H^{(l)}_{i})\ \ \textrm{if} \ \ l\neq k
\end{equation}
to stop back-propagation of the gradients through branches $l\neq k$ during the optimization. In other words, we prevent the data in language $k$ from affecting the optimization of other language branches, so that only the branch that matches the input batch language will be updated in a single iteration. Branch ${k}$ is effectively \textit {zero-shot} with respect to the languages $l\neq k$, 
 since branch ${k}$ has never been optimized on the data other than language $k$.
 
 We mix the detached representations
 \begin{equation}
(H^{(1)}_{i},\ldots )  = \textrm{Mixer}_{i}(\tilde H^{(1)}_{i},\ldots, \tilde H^{(N_{sp})}_{i})
\end{equation}
where the $\textrm{Mixer}_i$ performs the mixing at the $i$-th layer.
The outputs $(H^{(1)}_{i},\ldots )$ denote one or more representations fed to the next layer.

\subsubsection*{$\bullet$~~Mixer}
There are two types of mixers: the inner-mixers at layers $1\sim K-1$ and the outer-mixer at layer $K$. 
They are similar except that the inner-mixer has $N_{sp}$ outputs for $N_{sp}$ subsequent branches while the outer-mixer only has one output. In Fig \ref{fig:mixer} we demonstrate the structure of an inner-mixer, which mixes representations feature-wise with a simple attention-like mechanism. 
It contains $N_{sp}$ submodule.  Take the $M^{en}$ submodule for example. We take the dot product of a trainable vector $v^{en}\in \mathbb{R}^{n}$ with the inputs $\tilde H^{(1)},\ldots, \tilde H^{(N_{sp})}\in \mathbb{R}^{n\times d}$ and obtain the attention score vectors for each representations $h^{(1)}\ldots, h^{(N_{sp})} \in \mathbb{R}^{d}$. We stack the score vectors
\begin{equation}
h^{\text{stack}} = \textrm{Stack}(h^{(1)},\ldots, h^{(N_{sp})})\in \mathbb{R}^{N_{sp}\times d}
\end{equation}
and apply  softmax to get the mixing weight
 \begin{equation}
 \left[w^{(1)}; \ldots; w^{(N_{sp})}\right]  = \textrm{softmax}(h^{\text{stack}})
 \end{equation}
 The softmax is applied along the language dimension of size $N_{sp}$.
 $w^{(l)}\in \mathbb{R}^{d}$ measures the importance of each feature in $\tilde H^{(l)}$.
 The final output is given by the weighted sum of $w^{(l)}$ and $\tilde H^{(l)}$
\begin{equation}
H^{en} = \sum_{l}^{N_{sp}}\tilde H^{(l)}\odot w^{(l)} 
\end{equation}
where $\odot$ is the broadcasting element-wise product.

\begin{table}[t!]
    \small\centering
    \begin{tabular}{@{}lccccc@{}}
    \toprule
         & \#Language & \#Labels & \#Train & \#Dev & \#Test \\ \midrule
    XNLI & 15         & 3        & 392,702              & 2,490              & 5,010               \\
    MARC & 6          & 5        & 200,000              & 5,000              & 5,000               \\ \bottomrule
    \end{tabular}
    \caption{Statistics of XNLI and MARC datasets. Under \#Train, \#Dev and \#Test we show the number of examples per language.}
    \label{table:stats}
    \end{table}

\begin{table}[t!]
\small\centering

\begin{tabular}{@{}lcc@{}}
\toprule
\textbf{Teachers}              & \textbf{XNLI}  & \textbf{MARC}  \\ \midrule
(en) RoBERTa  & 91.08 & 67.66 \\
(fr) CamemBERT & 85.45 & 62.70 \\
(zh) RoBERTa-wwm-ext  & 80.86 & 60.00 \\ \midrule
Average       & 85.80 & 63.45 \\ \bottomrule
\end{tabular}

\caption{Accuracy scores ($\times$ 100\%) of the teachers on the XNLI and MARC test set.  The scores represent the performance of teachers on the corresponding languages.}
\label{table:monolingual_teachers}
\end{table}

\begin{table*}[t]
\resizebox{\linewidth}{!}{
\begin{tabular}{@{}lcccccccccccccccccc@{}}
\toprule
    & \multicolumn{3}{|c|}{\textbf{Supervised}}   & \multicolumn{12}{c|}{\textbf{Zero-shot}}&\multirow{2}{*}{{AVG(S)}}  &  \multirow{2}{*}{{AVG(Z)}} &  \multirow{2}{*}{{AVG(A)}} \\ \cmidrule(lr){2-16}

\multicolumn{1}{l|}{}        & en   & fr   &\multicolumn{1}{l|}{zh}      & de   & es   & ru   & ar   & hi   & vi   & bg   & el   & sw   & th   & tr   & \multicolumn{1}{l|}{ur}   &  &  &   \\ \midrule
\multicolumn{16}{@{}l|}{\textit{mBERT-based model}}  &        &        &        \\ 
{MonoKD} &84.29	&74.95	&71.00	&72.44	&75.88	&70.44	&66.00	&61.11	&71.78	&69.53	&67.44	&48.91	&52.59	&61.08	&\multicolumn{1}{l|}{58.83}	&76.75	&64.67	&67.08  \\
{MultiTrain}    &83.27	&78.53	&77.19	&73.88	&77.55	&72.85	&68.37	&64.92	&73.43	&72.44	&70.40	&51.27	&58.16	&65.15	&\multicolumn{1}{l|}{62.58}	&79.66	&67.58	&70.00  \\
{MultiKD}    &85.07	&79.57	&\textbf{78.65}	&74.58	&78.09	&73.40	&68.87	&\textbf{65.09}	&74.23	&72.86	&70.08	&51.10	&57.07	&64.83	&\multicolumn{1}{l|}{62.84}	&81.10	&67.75	&70.42  \\
{MultiKD + MBLM}   &\textbf{85.33}	&\textbf{79.73}	&\textbf{78.65}	&\textbf{75.22}	&\textbf{79.03}	&\textbf{73.65}	&\textbf{69.37}	&65.03	&\textbf{74.34}	&\textbf{73.46}	&\textbf{70.58}	&\textbf{51.81}	&\textbf{57.75}	&\textbf{64.98}	&\multicolumn{1}{l|}{\textbf{63.08}}	&\textbf{81.24}	&\textbf{68.19}	&\textbf{70.80}\\ \midrule
\multicolumn{16}{@{}l|}{\textit{XLM-R$_{\texttt{base}}$-based model}}   &        &        &        \\ 
{MonoKD}   &85.94	&79.57	&75.68	&78.40	&80.32	&77.09	&73.63	&72.02	&76.87	&79.32	&77.49	&66.27	&74.17	&74.42	&\multicolumn{1}{l|}{67.92}	&80.40	&74.83	&75.94   \\
{MultiTrain}     &85.45	&80.68	&78.95	&79.20	&80.91	&78.01	&75.05	&74.31	&78.07	&80.20	&78.11	&67.40	&75.84	&75.90	&\multicolumn{1}{l|}{70.20}	&81.69	&76.10	&77.22 \\
{MultiKD} &\textbf{86.70}	&\textbf{81.51}	&80.77	&79.73	&81.53	&78.78	&75.35	&74.80	&78.33	&81.46	&79.04	&67.08	&\textbf{76.88}	&76.18	&\multicolumn{1}{l|}{70.73}	&\textbf{82.99}	&76.66	&77.92 \\
{MultiKD + MBLM} &86.63	&\textbf{81.54}	&\textbf{80.87}	&\textbf{80.29}	&\textbf{82.14}	&\textbf{78.91}	&\textbf{75.61}	&\textbf{75.31}	&\textbf{78.82}	&\textbf{81.74}	&\textbf{79.18}	&\textbf{68.32}	&\textbf{76.93}	&\textbf{77.21}	&\multicolumn{1}{l|}{\textbf{71.49}}	&\textbf{83.01}	&\textbf{77.16}	&\textbf{78.33}     \\ \bottomrule          

\end{tabular}
}
\caption{Accuracy scores ($\times$ 100\%) on XNLI test set. \textit{AVG(S)} is the average scores on the three supervised languages
(strictly speaking, French and Chinese are not supervised languages for {monoKD}).
\textit{AVG(Z)} is the average scores on all the languages excluding English, French and Chinese. \textit{AVG(A)} is the average scores over all languages.  
We \textbf{bold} any score that is within $0.05$ of the best in a given experiment.
Results on dev set are listed in the appendix.} 
\label{table:main_results}
\end{table*}

\begin{table*}[t]
\centering
\resizebox{0.8\linewidth}{!}{

\begin{tabular}{@{}lcccccc|ccc@{}}
\toprule
                              & \multicolumn{3}{|c|}{\textbf{Supervised}}   & \multicolumn{3}{c|}{\textbf{Zero-shot}}&\multirow{2}{*}{{AVG(S)}}  &  \multirow{2}{*}{{AVG(Z)}} &  \multirow{2}{*}{{AVG(A)}} \\ \cmidrule(lr){2-7}

                              & \multicolumn{1}{|c}{en} & \multicolumn{1}{c}{fr} & \multicolumn{1}{c|}{zh} & de & \multicolumn{1}{c}{es}& \multicolumn{1}{c|}{ja} &   &  &   \\ \midrule
\textit{mBERT-based model}   &    &                        &                        &    &    &                         &                            &                            &                            \\
{MonoKD}                     &65.61	&51.11	&42.86	&52.97	&52.14	&39.36                   & 53.19                     & 48.16                      & 50.67                      \\
{MultiTrain}          &65.83	&61.22	&57.71	&54.38	&54.81	&41.48                 & 61.59                      & 50.22                      & 55.90                      \\
{MultiKD}                        &{66.22}	&\textbf{61.52}	&\textbf{58.62	} &55.35	&55.54	&41.54             & \textbf{62.12}                      & 50.81                      & 56.46                      \\
{MultiKD + MBLM}   &\textbf{66.29}	&\textbf{61.57} 	&\textbf{58.59	} &\textbf{56.57}	&\textbf{56.87	}&\textbf{42.43}                    & \textbf{62.15}                      & \textbf{51.96}                      & \textbf{57.05}                     \\ \midrule
\textit{XLM-R$_{\texttt{base}}$-based model}   &    &                        &                        &    &    &                         &                            &                            &                            \\
{MonoKD}            &66.61	&58.30	&52.83	&64.22	&58.39	&\textbf{54.87}	      & 59.25                      & 59.16                      & 59.20                      \\ 
{MultiTrain}     &66.60	&61.62	&59.13	&63.98	&60.38	&50.97                    & 62.43                      & 58.46                      & 60.45                      \\ 
{MultiKD}           & \textbf{66.90}	&\textbf{62.18} &\textbf{59.61}	&64.82	&60.87	&50.59	                & \textbf{62.90}                      & 58.76                     & 60.83                      \\
{MultiKD + MBLM}   &\textbf{66.92}	&\textbf{62.25}	&59.43	&\textbf{65.67} &\textbf{61.04}	&51.17                 & \textbf{62.87}                      & \textbf{59.30}                      & \textbf{61.08}                      \\ \bottomrule
\end{tabular}
}
\caption{Accuracy scores ($\times$ 100\%) on MARC test set.} 

\label{table:main_results_pawsx}
\end{table*}

\section{Experiments}
\subsection{Datasets}

\noindent \textbf{XNLI}. The Cross-Lingual Natural Language Inference (XNLI) \cite{Conneau2018XNLIEC} asks whether a premise sentence entails, contradicts, or is neutral toward a hypothesis sentence.
It covers 15 languages. The English dev and test sets are translated into other languages by professional translators. 
For training, we use the machine-translated data provided by the authors.

\noindent \textbf{MARC}. Multilingual Amazon Reviews Corpus (MARC) \cite{keung-etal-2020-multilingual} is a large-scale collection of Amazon reviews for multilingual text classification. It covers 6 languages. Each example is labeled by one of 5 possible star ratings (from 1 to 5). A major difference between MARC and XNLI is that MARC contains natural (non-translated) training and evaluation data for non-English languages.

The statistics of XNLI and MARC are listed in Table \ref{table:stats}.

\subsection{Experimental Setup}

We experiment with mBERT and XLM-R$_{\texttt{base}}$ as our student models.
We take English, French and Chinese as the supervised languages\footnote{We choose these languages because there are public available high-performance monolingual models for these languages.}.
For the monolingual teachers, we use  RoBERTa for English \cite{liu2019roberta},  CamemBERT for French \cite{martin-etal-2020-camembert} and RoBERT-wwm-ext for Chinese \cite{cui-etal-2020-revisiting}. 
All of the teachers are of the same structure with 24 layers, 16 attention heads, which are largely the same as BERT-large, except that they do not use token type embeddings.
All the {MBLM} models have three language branches, and the depth $K$ of each branch is $4$. All the branches are initialized with the same parameters by copying from the bottom four layers in mBERT or XLM-R.

\indent For both tasks, we search the learning rate form \{1e-5, 2e-5, 3e-5\} and number of epochs from \{2,3,5,7,10\}. The batch size is 32. The maximum length is $128$. Temperature is $8$ for all distillation experiments.
We select the best model based on its mean performance over all languages on the dev set. We run each experiment 5 times with different seeds and report the mean score. 
We implement the model with Transformers \cite{wolf-etal-2020-transformers}. We train and distill the models with TextBrewer \cite{textbrewer-acl2020-demo}.
All the experiments were performed with a single V100 GPU.

\subsection{Baselines}
\noindent{\textbf{MonoKD}}. Monolingual knowledge distillation. We take the English RoBERTa as the teacher, fine-tune it on the English training data,  then distill it to the multilingual student.

\noindent{\textbf{MultiTrain}}. Multilingual Training. We directly train the multilingual model with all supervised training data (English, French and Chinese).

\subsection{Results}

\begin{figure*}[t!]
\centering
\includegraphics[width=.95\linewidth]{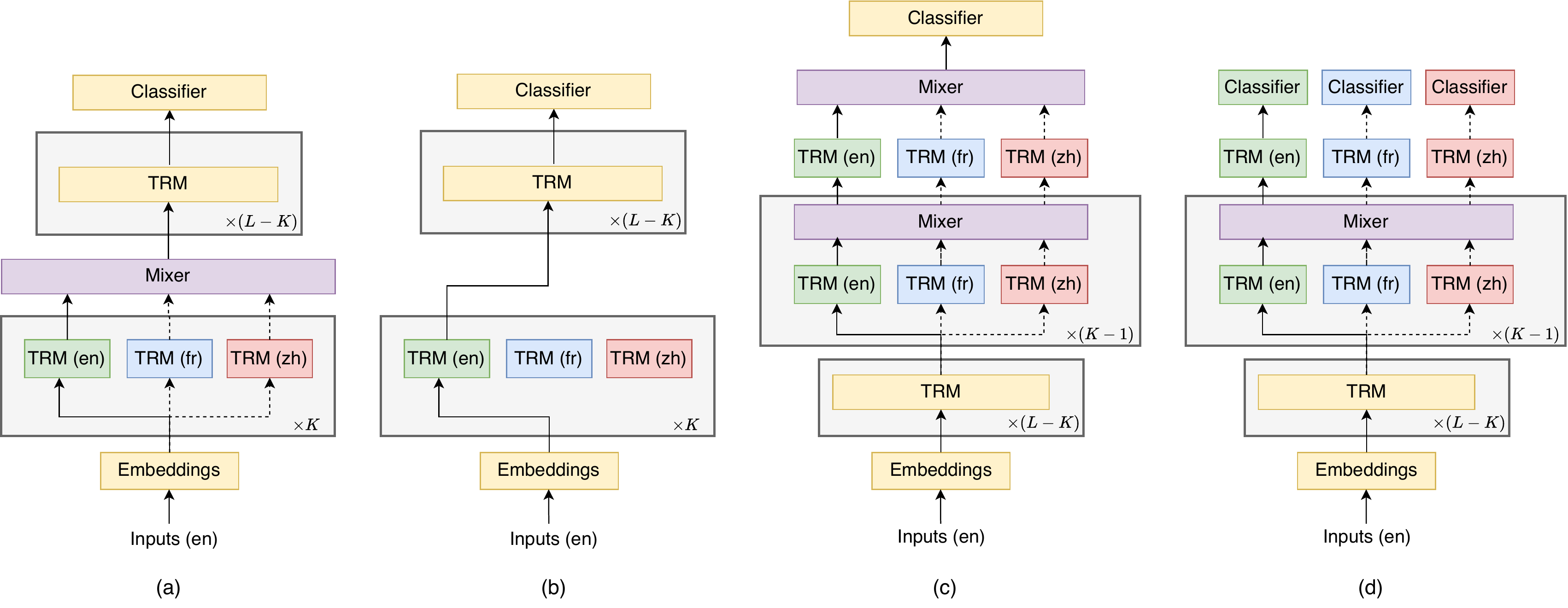}
\caption{Different model structures in the ablation studies. (a) No intermediate mixers. (b) Remove all mixers. (c) Language branches are placed after shared encoders, with a single classifier head. (d) Language branches are placed after shared encoders, with multiple classifier heads.}
\label{fig:ablations}
\end{figure*}

For reference, we list the performance of monolingual teachers in Table \ref{table:monolingual_teachers},  which sets the approximate upper bound performance of the student model on supervised languages.

We compare our method and the baselines on XNLI  in Table \ref{table:main_results}.
Comparing {MonoKD} and multilingual distillation ({MultiKD}), we can conclude that distillation with more languages notably improves both supervised performance and zero-shot performance.
Comparing {MultiTrain} and {MultiKD}, we see that the supervised performance is improved more significantly than zero-shot performance.
By replacing the standard transformer blocks in the student model with {MBLM}, {MultiKD+MBLM} achieves the best performance. Especially, the performance on zero-shot languages gets improved more notably than supervised performance, which is complementary to {MultiKD}.

The results on MARC are listed in Table \ref{table:main_results_pawsx}, with the classification accuracy as the metric
\footnote{In \citet{keung-etal-2020-multilingual}, authors suggest using mean absolute error (MAE) as the primary metric. Here we use classification accuracy for consistency with XNLI. The results measured in MAE are shown in the appendix.}.
Compared with the {MultiTrain} baseline,  {MultiKD} improves the supervised performance.
Replacing the student with {MBLM} further improves the zero-shot performance. This effect is more evident on MARC than XNLI. 
Finally, the combination of the techniques {MultiKD+MBLM} achieves the best performance on this task.

\begin{table}[t]
\resizebox{\linewidth}{!}{
\begin{tabular}{@{}lccc@{}}
\toprule
                                                & AVG(S) & AVG(Z) & AVG(A) \\ \midrule
\multicolumn{1}{l}{mBERT}              &81.63	&67.53	&70.35     \\
\multicolumn{1}{l}{\ + {MBLM}}             & \textbf{81.87}  & \textbf{68.35}  & \textbf{71.06}     \\ \midrule
\multicolumn{1}{l}{\ \ \ - inner-mixers}     & 81.59      & 68.00      & 70.72      \\
\multicolumn{1}{l}{\ \ \ - all mixers (single)}     &81.69	 & 67.84	&70.61     \\
\multicolumn{1}{l}{\ \ \ - all mixers (all)}        &  81.30      & 67.84      & 70.53     \\ 
\multicolumn{1}{l}{\ \ \ -  detaching} & 81.66    & 67.37     & 70.23     \\ \midrule
\multicolumn{1}{l}{\ +  branches at the top}     & 79.48      & 65.09      & 67.97      \\
\multicolumn{1}{l}{\ \ \ +  multi-classifier} & 81.57     & 67.53     & 70.34     \\ 
\bottomrule
\end{tabular}
}
\caption{Ablation studies on XNLI dev set. The experiments are performed in the {MultiKD} setting.}
\label{table:ablation}
\end{table}
\subsection{Ablation Studies}

In the {MBLM}, is it necessary to mix the representations from other branches? Why mixing the representations at the bottom layers?
We conduct two groups of experiments to answer these questions. The first group is to study the necessity of mixing and zero-shot-aware training:
\begin{itemize}[leftmargin=*,noitemsep]
\item \textbf{Without inner-mixers} (Fig \ref{fig:ablations} (a)). We remove the inner-mixers and only keep the outer-mixer. The representations from different branches only mix once at the 4th layer.
\item \textbf{Without all mixers} (Fig \ref{fig:ablations} (b)).  We remove all the mixers between language branches. During training, each batch only goes through the corresponding language branch. During inference, for zero-shot languages, the batch goes through all branches, and we average the representations; for supervised languages, we have two options: either feed the batch only to the corresponding language branch (\textbf{single}), or feed the batch to all branches and average the representations (\textbf{all}).
\item \textbf{Without detaching}. During training, we remove the detach mechanism, i.e., we skip Eq \eqref{eq:detach}, so that the gradients can backpropagate freely through all the branches.
\end{itemize}
The second group is to study the effects of replacing transformers with language branches at different layers.
We experiment with model structures that share bottom layers between languages:
\begin{itemize}[leftmargin=*,noitemsep]
\item Language branches at the top (\textbf{branches at the top} (Fig \ref{fig:ablations} (c)). Instead of sharing the top layers between different languages, the bottom layers are shared.
the language branches are built on top of the shared encoders.
\item Language branches at the top with \textbf{multi-classifier} (Fig \ref{fig:ablations} (d)), which is largely the same as the structure above, except that instead of sharing the classifier head, 
each language branch has its classifier head. For zero-shot languages, we average the logits from all the classifier heads.
\end{itemize}

We perform the ablations in the {MultiKD} setting.
The results are shown in Table \ref{table:ablation}. 
The first group of experiments shows the importance of mixing representations at different layers and zero-shot-aware training. Both the outer-mixer and inner-mixers are crucial to the performance. Especially, when all the mixers are removed, the representations mix only at the inference stage, and the results are worse. 
On the other hand, if we keep the MBLM structure unchanged, but train it in the normal way, without the detach mechanism (the experiment "-detaching"),  we see that the performance is nearly the same as the mBERT.
Thus above results prove the necessity of training stage representation mixing and zero-shot-aware training.

The results of the second group show that simply moving language branches to the top layers degrades the performance. It may be because that there are no following transformers to extract the features from the mixed representations. Changing the output classifier to the multi-classifier relieves the problem, but there are no obvious benefits compared with the mBERT.

\section{Analysis}
\subsection{Depth of Language Branches}
\begin{figure}[tbp]
\centering
\includegraphics[width=\columnwidth]{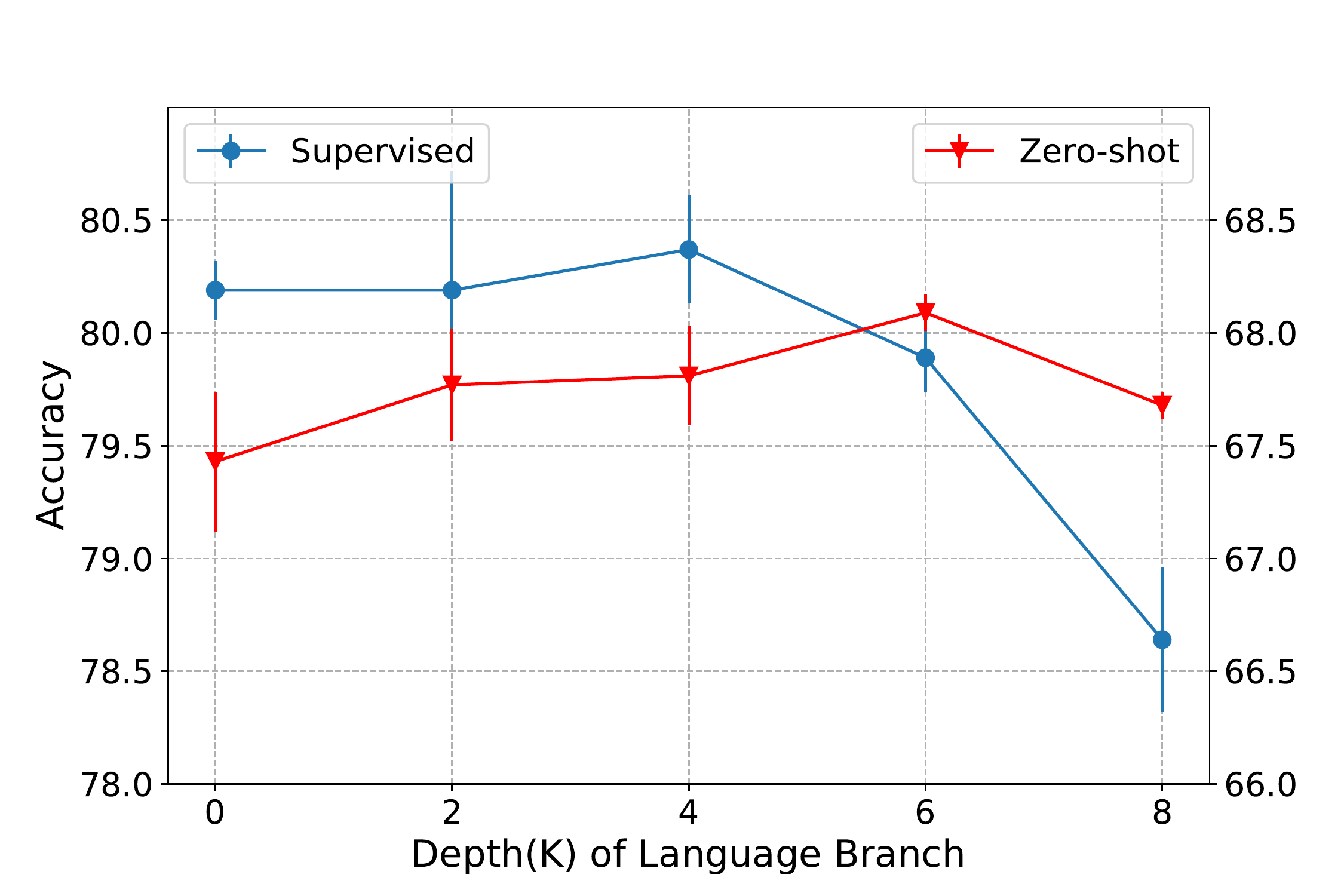}
\caption{The accuracy scores of models with different language branch depths on XNLI dev set. Left y-axis: supervised accuracy. Right y-axis: zero-shot accuracy.}
\label{fig:depth}
\end{figure}

In the previous experiments, we have fixed the depth of the language branch to $K=4$.
In this section, we fix the total number of layers to 12 and investigate the effect of varying depth of language branches.
As the language branches become deeper, the model grows bigger and slower for training and inference.
For simplicity, we perform normal training rather than distillation on the mBERT-based {MBLM}. We train all the models with the same learning rate $1e-5$ and the number of epochs $3$. We vary the depths of language branches and plot the accuracy on XNLI in Fig \ref{fig:depth}.

As the language branches grow deeper, the zero-shot performance peaks around $K=6$, and the supervised performance peaks around $K=4$. But the supervised performance drops notably as $K$ keeps growing. Thus, we choose $K=4$ in our main experiments to keep a balanced performance and decent model size.
Besides, since $K=0$ represents the standard mBERT model, the plot also shows that the MBLM structure is effective not only in the distillation but normal training.

\subsection{Model Efficiency}
Compared with standard transformer models, the MBLM introduces 8 additional transformer blocks and 4 mixer layers,
which contains approximately $32\%$ more parameters compared with mBERT ($20\%$ more parameters compared with XLM-R$_{\texttt{base}}$). Could the improvements barely come from the increase in the model size? To answer this question, we compare the models with different numbers of parameters.
The results are listed in Table \ref{table:model_efficiency}. For the 2-model-ensemble (Ens$_{2}$), we average the predicted logits from two mBERT models.
Ens$_{2}$ outperforms the {MBLM} on supervised languages, but on zero-shot languages, Ens$_{2}$ is worse than the {MBLM}, which contains fewer parameters. The result suggests that the improvement is not merely a consequence of enlarged model capacity.
During training, since only one language branch is trained in each iteration, the additional parameters do not significantly slow down the training process of MBLM.
The total training time is only $1.1$ times slower than the single model.

\begin{table}[t]
\resizebox{\linewidth}{!}{
\begin{tabular}{@{}lcccll@{}}
\toprule
    & AVG(S) & AVG(Z) & AVG(A)   & Params     & Time       \\ \midrule
\multicolumn{5}{l}{\textit{mBERT-based model}}       &         \\ 
\multicolumn{1}{l}{Single}  & 81.63 & 67.53 & 70.35 & 1$\times$   & 1$\times$ \\
\multicolumn{1}{l}{Ens$_2$}   & \textbf{82.06}     & 67.95      & 70.80     & 2$\times$   & 2$\times$ \\
\multicolumn{1}{l}{{MBLM}} &81.87     & \textbf{68.35}     & \textbf{71.06}     & 1.3$\times$ & 1.1$\times$ \\ \bottomrule
\end{tabular}
}
\caption{Performance (accuracy) of models with different numbers of parameters on XNLI dev set. \textit{Params} is the number of model parameters. 
\textit{Time} is the training time.
 \textit{Ens}$_2$ indicates a 2-model-ensemble. 
 The experiments are performed in the {MultiKD} setting.}
\label{table:model_efficiency}
\end{table}

\section{Conclusion}
In this work, we have studied how to improve multilingual model performance for both supervised and zero-shot languages with only resources from supervised languages available. We combined multilingual knowledge distillation ({MultiKD}) and the proposed {MBLM} structure, together with the zero-shot-aware training strategy. The key idea is mixing the zero-shot representations from multiple language branches during training. We tested our method on two text classification tasks. 

Experimentally we find that {MultiKD} and {MBLM} are complementary to each other.
With {MultiKD}, the student model learns from multiple monolingual teachers and improves its performance, especially on supervised languages. By replacing the standard transformer model with the {MBLM}, the zero-shot cross-lingual transferability is notably improved. Further analysis justifies the model structure and zero-shot-aware training strategy.

\bibliography{main} 

\appendix
\section{Data Processing}

\label{sec:appendix}

For XNLI, we concatenate the premise and the hypothesis of each example and use pre-trained language models for sentence pair classification. 
For MARC, input is the review body combined with the review title and product category. We treat the review body as the first sentence, the review title concatenated with the product category as the second sentence. We use the same sentence pair classification model as XNLI.

\section{More Results}

\begin{table*}[thbp]
\resizebox{\linewidth}{!}{
\begin{tabular}{@{}lcccccccccccccccccc@{}}
\toprule
    & \multicolumn{3}{|c|}{\textbf{Supervised}}   & \multicolumn{12}{c|}{\textbf{Zero-shot}}&\multirow{2}{*}{{AVG(S)}}  &  \multirow{2}{*}{{AVG(Z)}} &  \multirow{2}{*}{{AVG(A)}} \\ \cmidrule(lr){2-16}

\multicolumn{1}{l|}{}        & en   & fr   &\multicolumn{1}{l|}{zh}      & de   & es   & ru   & ar   & hi   & vi   & bg   & el   & sw   & th   & tr   & \multicolumn{1}{l|}{ur}   &  &  &   \\ \midrule
\multicolumn{4}{l}{\textit{mBERT-based model}}       &      &      &      &      &      &      &      &      &      &      &       & \multicolumn{1}{l|}{}  &        &        &        \\ 
\textsc{MonoKD}  &84.81	&75.01	&70.88	&73.43	&76.46	&68.75	&65.73	&61.28	&71.55	&68.69	&66.59	&49.01	&51.47	&62.09	&\multicolumn{1}{l|}{58.88} & 76.90     & 64.49   & 66.98   \\
\textsc{MultiTrain}    &83.10	&79.11	&78.34	&74.66	&78.13	&70.71	&68.53	&65.19	&73.47	&71.38	&69.26	&53.33	&56.86	&65.81	&\multicolumn{1}{l|}{61.76}    & 80.18  & 67.43  & 69.98  \\
\textsc{MultiKD}    &85.10	&\textbf{80.42	}&79.35	&75.33	&79.06	&71.22	&68.62 &65.16	&73.44	&71.34	&69.78	&52.54	&56.58	&65.23	 &\multicolumn{1}{l|}{62.08}     & 81.63  & 67.53  & 70.35  \\
\textsc{MultiKD + MBLM}   &\textbf{85.31}	&\textbf{80.46	}&\textbf{79.84}	&\textbf{76.14}	&\textbf{79.82} 	&\textbf{71.94} 	&\textbf{68.71	}&\textbf{65.93}	&\textbf{74.71}	&\textbf{72.84}	&\textbf{69.88}	&\textbf{53.38}	&\textbf{58.15}	&\textbf{66.39}	&\multicolumn{1}{l|}{\textbf{62.31}}     & \textbf{81.87} & \textbf{68.35}  & \textbf{71.06}  \\ \midrule
\multicolumn{4}{l}{\textit{XLM-R$_{\texttt{base}}$-based model}}       &      &      &      &      &      &      &      &      &      &      &       & \multicolumn{1}{l|}{}  &        &        &        \\ 
\textsc{MonoKD}    &85.66	&79.71	&76.39	&79.23	&80.83	&77.04	&73.14	&71.97	&76.47	&78.59	&77.60	&65.53	&74.30	&74.93	&\multicolumn{1}{l|}{68.42} & 80.59  & 74.84   & 75.99   \\
\textsc{MultiTrain}      &85.32	&80.84	&79.33	&79.98	&81.78	&77.65	&74.53	&73.96	&77.65	&79.31	&78.35	&65.89	&76.21	&76.27	&\multicolumn{1}{l|}{70.46}    & 81.83  & 76.00  & 77.17  \\ 
\textsc{MultiKD} &87.41	&81.95	&\textbf{81.50}	&81.17	&82.72	&78.90	&\textbf{75.36}	&74.26	&79.29	&80.90	&79.12	&66.11	&77.20	&77.21	&\multicolumn{1}{l|}{70.89}     & 83.62  & 76.93  & 78.27  \\
\textsc{MultiKD + MBLM} &\textbf{87.72}	&\textbf{82.37}	&\textbf{81.54}	&\textbf{81.39}	&\textbf{83.45}	&\textbf{79.19}	&\textbf{75.39}	&\textbf{75.42}	&\textbf{79.89}	&\textbf{81.11}	&\textbf{79.64}	&\textbf{67.27}	&\textbf{78.07}	&\textbf{77.56}	&\multicolumn{1}{l|}{\textbf{71.74}}     & \textbf{83.88}        & \textbf{77.51}       & \textbf{78.78}     \\ \bottomrule          

\end{tabular}
}
\caption{Accuracy scores ($\times$ 100\%) on XNLI dev dataset. }
\label{table:xnli_results_dev}
\end{table*}

\begin{table*}[t]
\centering
\resizebox{0.8\linewidth}{!}{

\begin{tabular}{@{}lcccccc|ccc@{}}
\toprule
                              & \multicolumn{3}{|c|}{\textbf{Supervised}}   & \multicolumn{3}{c|}{\textbf{Zero-shot}}&\multirow{2}{*}{{AVG(S)}}  &  \multirow{2}{*}{{AVG(Z)}} &  \multirow{2}{*}{{AVG(A)}} \\ \cmidrule(lr){2-7}

                              & \multicolumn{1}{|c}{en} & \multicolumn{1}{c}{fr} & \multicolumn{1}{c|}{zh} & de & \multicolumn{1}{c}{es}& \multicolumn{1}{c|}{ja} &   &  &   \\ \midrule
\textit{mBERT-based model}   &    &                        &                        &    &    &                         &                            &                            &                            \\
{MonoKD}                     &39.73	&62.71	&80.75	&57.83	&59.99	&80.89                   & 61.06                     & 66.23                      & 63.65                      \\
{MultiTrain}          &39.49	&43.50	&52.56	&55.79	&53.54	&77.68                 & 45.18                      & 62.34                      & 53.76                      \\
{MultiKD}                        &{39.02}	&{42.85}	&{51.68	} &54.17	&53.07	&77.47             & {44.52}                      & 61.57                      & 53.04                      \\
{MultiKD + MBLM}   &\textbf{38.70}	&\textbf{42.38} 	&\textbf{51.02} &\textbf{51.51}	&\textbf{50.92}&\textbf{76.71}                    & \textbf{44.03}                      & \textbf{59.71}                      & \textbf{51.87}                     \\ \bottomrule
\end{tabular}
}
\caption{MAE scores ($\times$ 100\%) of mBERT-based model on MARC test set. MAE scores are the lower the better.} 

\label{table:marc_results_mae}
\end{table*}

In Table \ref{table:xnli_results_dev} we shown the accuracy scores on XNLI dev set.
In Table \ref{table:marc_results_mae} we show the mean absolute error (MAE) scores of mBERT-based model on the  MARC test set.

\end{document}